\pdfoutput=1

\documentclass[11pt]{article}

\usepackage[]{emnlp2021}

\usepackage{times}
\usepackage{latexsym}
\usepackage{graphicx}
\usepackage{multirow}
\usepackage{booktabs}
\usepackage{amssymb,amsfonts,amsmath}

\usepackage[T1]{fontenc}

\usepackage[utf8]{inputenc}

\usepackage{microtype}

\title{Predicting emergent linguistic compositions through time: \\ Syntactic frame extension via multimodal chaining}

\author{
  Lei Yu$^{1}$, Yang Xu$^{1,\,2,\,3}$ \\
  $^1$ Department of Computer Science, University of Toronto, Toronto, Canada \\
  $^2$ Cognitive Science Program, University of Toronto, Toronto, Canada \\
  $^3$ Vector Institute for Artificial Intelligence, Toronto, Canada \\
  \texttt{\{jadeleiyu,yangxu\}@cs.toronto.edu}
}
\begin{document}
\maketitle
\begin{abstract}
Natural language relies on a finite lexicon to express an unbounded set of emerging ideas. One result of this tension is the formation of new compositions, such that existing linguistic units can be combined with emerging items into novel expressions. We develop a framework that exploits the cognitive mechanisms of chaining and multimodal knowledge to predict emergent compositional expressions through time. We present the syntactic frame extension model (SFEM) that draws on the theory of chaining and knowledge from ``percept'', ``concept'', and ``language'' to infer how verbs extend their frames to form new compositions with existing and novel nouns. We evaluate SFEM rigorously on the 1) modalities of knowledge and 2) categorization models of chaining, in a syntactically parsed  English corpus over the past 150 years. We show that multimodal SFEM
predicts newly emerged verb syntax and arguments substantially better
than competing models using purely linguistic or unimodal knowledge. We  find support for an exemplar view of chaining as opposed to a prototype view and reveal how the joint approach of multimodal chaining may be fundamental to the creation of literal and figurative language uses including metaphor and metonymy.
\end{abstract}

\section{Introduction}

Language users often construct novel compositions through time, such that existing linguistic units can be combined with emerging items to form novel expressions.  Consider the expression {\it swipe your phone}, which presumably came about after the emergence of touchscreen-enabled smartphones. Here the use of the verb {\it swipe} was extended to express one’s experience with the emerging item ``smartphone''. These incremental extensions are fundamental to adapting a finite lexicon toward emerging communicative needs. We explore the nature of cognitive mechanisms and knowledge in the temporal formation of previously unattested verb-argument compositions, and how this compositionality  may be understood in principled terms.

\begin{figure*}[t]
\centering
\includegraphics[width=1\textwidth]{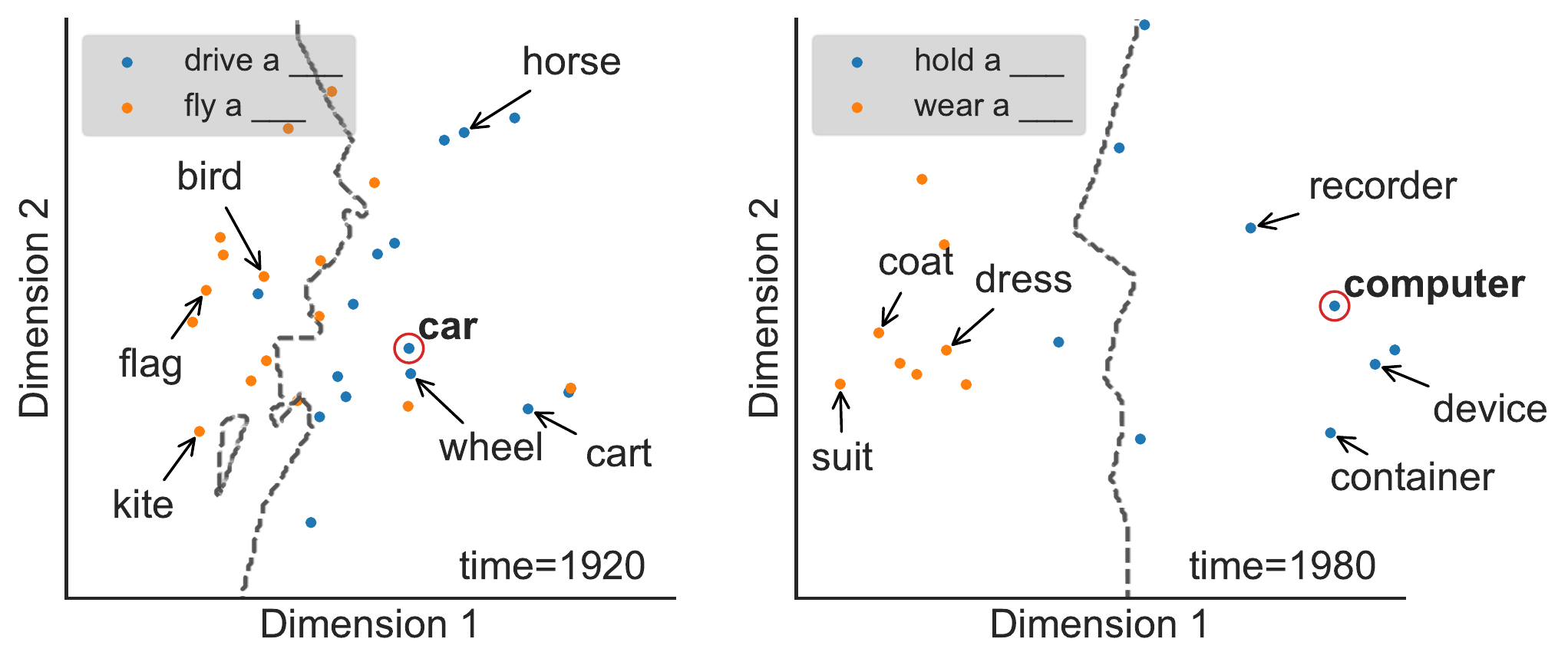}
\caption{Preview of the proposed approach of syntactic frame extension. Given a query noun (circled dot) at time $t$, the framework draws on a combination of multimodal knowledge and cognitive mechanisms of chaining to predict novel linguistic expressions for those items. The left panel shows via PCA projection how a newly emerged noun (i.e., \textit{car} in 1920s) is assigned appropriate verb frames by comparing the learned multimodal representation of the query nouns with support nouns (non-circled dots) that have been predicated by the frames. The right panel shows a similar verb construction for a noun that already existed at the time of prediction (i.e., \textit{computer} in 1980s).}
\label{noun_extensions}
\end{figure*}

Compositionality is at the heart of linguistic creativity yet a notoriously challenging topic in computational linguistics and natural language processing (e.g.,~\citealt{vecchi2017spicy,cordeiro2016predicting,blacoe2012comparison,mitchell2010composition,baroni2010nouns}). For instance, modern views on the state-of-the-art neural models of language have suggested that  they show some degree of linguistic generalization but are impoverished in systematic compositionality (see \citet{baroni2020linguistic} for review). Existing work has also explored the efficacy of neural models in modeling diachronic semantics (e.g., \citealp{hamilton2016diachronic,rosenfeld2018deep,hu2019diachronic,giulianelli-etal-2020-analysing}). However, to our knowledge, no attempt has been made to examine principles in the formation of novel verb-noun compositions through time.


We formulate the problem as an inferential process which we call \emph{ syntactic frame extension}.  We define syntactic frame as a joint distribution over a verb predicate, its noun arguments, and their syntactic relations, and we focus on tackling two related predictive problems: 1) given a novel or existing noun, infer what verbs and syntactic relations that have not predicated the noun might emerge to describe it over time (e.g., {\it to \underline{drive} a car} vs. {\it to \underline{fly} a car}), and 2) given a verb predicate and a syntactic relation, infer what nouns can be plausibly introduced as its novel arguments in the future (e.g., {\it drive a \underline{car}} vs. {\it drive a \underline{computer}}). 

Figure \ref{noun_extensions} offers a preview of our framework by visualizing the process of assigning novel verb frames to describe two query nouns over time. In the first case, the model incorporated with perceptual and conceptual knowledge successfully predicts the verb  {\it drive} to be a better predicate than {\it fly} for describing the novel item \textit{car} that just emerged at the time of prediction where linguistic usages are not yet observed (i.e., emergent verb composition with a novel noun concept). In the second case, the model predicts that {\it hold} is a better predicate than {\it wear} for describing the noun \textit{computer}, which already existed at the time of prediction (i.e., emergent verb composition with an existing noun). 

Our approach connects two strands of research that were rarely in contact: cognitive linguistic theories of chaining and computational representations of multimodal semantics. Work in the cognitive linguistics tradition has suggested that frame extension is not arbitrary and involves the comparison between a new item to existing items that are relevant to the frame~\cite{fillmore1985frame}. Similar proposals lead to the theory of chaining postulating that linguistic categories grow by linking novel referents to existing ones of a word due to proximity in semantic space \cite{lakoff1987women, malt1999knowing,xu2016historical,ramiro2018algorithms,habibi2020chaining,grewal2020chaining}. However, such a theory has neither been formalized nor evaluated to predicting verb frame extensions through time. Separately, computational work in multimodal semantics has suggested how word meanings warrant a richer representation beyond purely linguistic knowledge (e.g.,~\citealt{bruni2012distributional,gella-etal-2016-unsupervised,gella2017disambiguating}). However, multimodal semantic representations have neither been examined in the diachronics of compositionality nor in light of the cognitive theories of chaining. We show that a unified framework that incorporates the cognitive mechanisms of chaining through deep models of categorization and multimodal semantic representations  predicts the temporal emergence of novel noun-verb compositions.



\section{Related work}

Our work synthesizes the interdisciplinary areas of cognitive linguistics, diachronic semantics, meaning representation, and deep learning.

\subsection{Cognitive mechanisms of chaining}

The problem of syntactic frame extension concerns the cognitive theory of chaining~\cite{lakoff1987women,malt1999knowing}. It has been proposed that the historical growth of linguistic categories depends on a process of chaining, whereby novel items link to existing referents of a word that are close in semantic space, resulting in chain-like structures. Recent studies have formulated chaining as models of categorization from classic work in cognitive science. Specifically, it has been shown that chaining may be formalized as an exemplar-based mechanism of categorization emphasizing semantic neighborhood profile~\cite{nosofsky1986attention}, which contrasts with a prototype-based mechanism that emphasizes category centrality~\cite{reed1972pattern,lakoff1987women,rosch1975cognitive}. This computational approach to chaining has been applied to explain word meaning growth in numeral classifiers~\cite{habibi2020chaining} and adjectives~\cite{grewal2020chaining}. Unlike these previous studies, we consider the open issue whether cognitive mechanisms of chaining might be generalized to verb frame extension which draws on rich sources of knowledge. It remains critically undetermined how ``shallow models'' such as the exemplar model can function or integrate with deep neural models~\cite{mahowald2020counts,mcclelland2020exemplar}, and how it might fair with the alternative  mechanism of prototype-based chaining in the context of verb frame extension. We address both of these theoretical issues in a framework that explores these alternative mechanisms of chaining in light of probabilistic deep categorization models.


\subsection{Diachronic semantics in NLP}

The recent surge of interest in NLP on diachronic semantics has developed Bayesian models of semantic change (e.g., \citealp{frermann2016bayesian}), diachronic word embeddings (e.g., \citealp{hamilton2016diachronic}), and deep contextualized language models (e.g., \citealp{rosenfeld2018deep,hu2019diachronic,giulianelli-etal-2020-analysing}). A common assumption in these  studies is that linguistic usages (from historical corpora) are sufficient to capture diachronic word meanings. However, previous work has suggested that text-derived distributed representations tend to miss important aspects of word meaning, including perceptual features \cite{andrews2009integrating, baroni2008concepts, baroni2010strudel} and relational information \cite{neccsulescu2015reading}. It has also been shown that both relational and perceptual knowledge are essential to construct creative or figurative language use such as metaphor \cite{gibbs2004metaphor, gentner2008metaphor} and metonymy \cite{radden1999towards}. Our work examines the function of multimodal semantic representations in capturing diachronic verb-noun compositions, and the extent to which such representations can be integrated with the cognitive mechanisms of chaining.

\subsection{Multimodal representation of meaning}

Computational research has shown the effectiveness of grounding language learning and distributional semantic models in multimodal knowledge beyond linguistic knowledge \cite{lazaridou2015combining,hermann2017grounded}. For instance, \citet{kiros2014unifying} proposed a pipeline that combines image-text embedding models with LSTM neural language models. \citet{bruni2014multimodal} identifies discrete ``visual words'' in images, so that the distributional representation of a word can be extended to encompass its co-occurrence with the visual words of images it is associated with. \citet{gella2017disambiguating} also showed how visual and multimodal information help to disambiguate verb meanings. Our framework extends these studies by incorporating the dimension of time into exploring how multimodal knowledge  predicts novel language use. 

\subsection{Memory-augmented deep learning}

Our framework also builds upon recent work on memory-augmented deep learning \cite{vinyals2016matching,snell2017prototypical}.  In particular, it has been shown that category representations enriched by deep neural networks can effectively generalize to few-shot predictions with sparse input, hence yielding human-like abilities in classifying visual and textual data \cite{pahde2020multimodal, singh2020end, holla2020learning}. In our work, we consider the scenario of constructing novel compositions as they emerge over time, where sparse linguistic information is  available. We therefore extend the existing line of research to investigate how representations learned from naturalistic stimuli (e.g., images) and structured knowledge (e.g., knowledge graphs) can reliably model the emergence of flexible language use that expresses new knowledge and experience.

\begin{figure*}[t]
\centering
\includegraphics[width=1\textwidth]{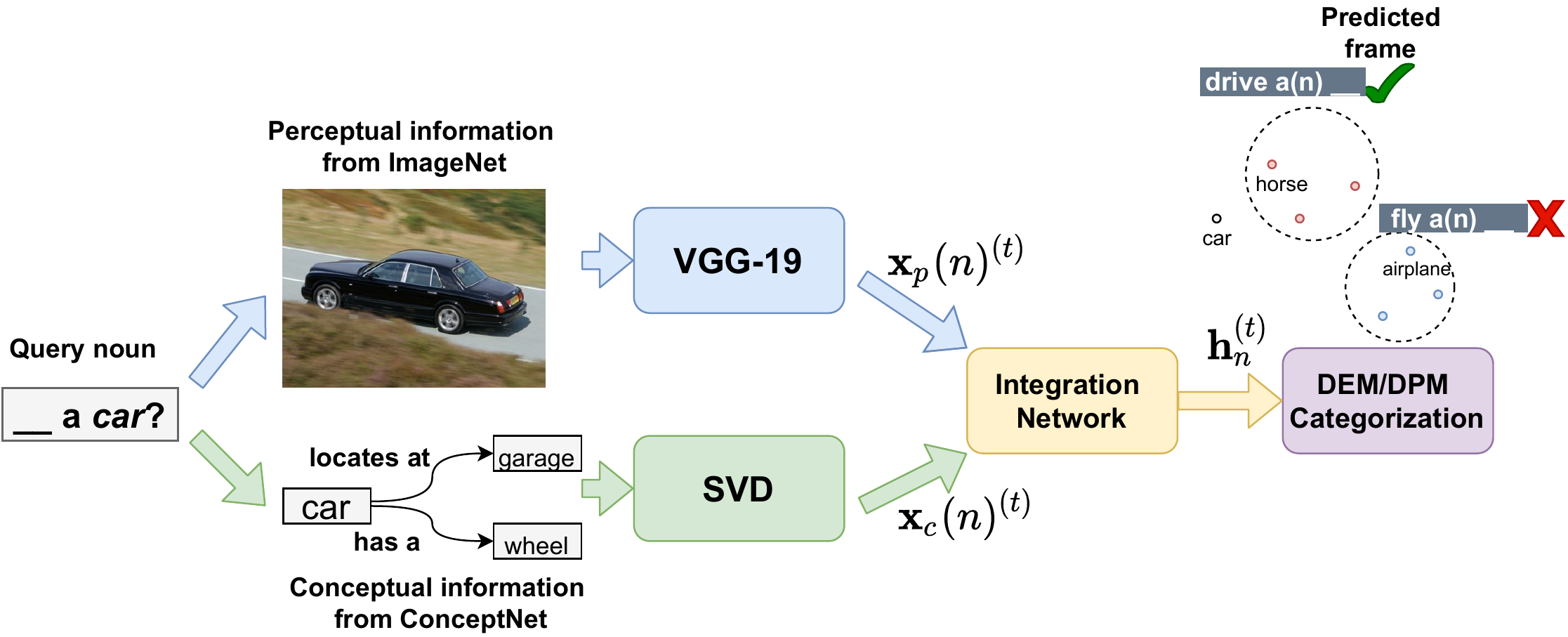}
\caption{Illustration of the syntactic frame extension model for the emerging query \textit{car}. The model integrates information from visual perception and conceptual knowledge about cars to form a multimodal embedding ($h_n^{(t)}$), which supports temporal prediction of appropriate verb-syntax usages via deep categorization models of chaining.}
\label{sfem_diagram}
\end{figure*}

\section{Computational framework}

We present the syntactic frame extension model (SFEM), which is composed of two components. First, SFEM specifies a frame as a joint probabilistic distribution over a verb, its noun arguments, and their syntactic relations and supports temporal prediction of verb syntax and arguments via deep probabilistic models of categorization. Second, SFEM draws on multimodal knowledge by incorporating perceptual, conceptual, and linguistic cues into flexible inference for extended verb frames over time. Figure \ref{sfem_diagram} illustrates our framework.

\subsection{Chaining as  probabilistic  categorization} \label{prob}

We denote a predicate verb as $v$ (e.g., \textit{drive}) and a syntactic relation as $r$ (e.g., direct object of a verb), and consider a finite set of verb-syntactic frame elements $f = (v, r) \in \mathcal{F}$. We define the set of nouns that appeared as arguments for a verb (under historically attested syntactic relations) up to time $t$ as support nouns, denoted by $n_s \in S(f)^{(t)}$ (e.g., \textit{horse} appeared as a support noun---the direct object---for the verb \textit{drive} prior to 1880s). 
Given a query noun $n^*$ (e.g., {\it car} upon its emergence in 1880s) that has never been an argument of $v$ under relation $r$, we define syntactic frame extension as probabilistic inference in two related problems:

\begin{itemize}
    \item[1.] {\bf Verb syntax prediction.} Here we predict which verb-syntactic frames $f$ are appropriate to describe the query noun $n^*$, operationalized as $p(f|n^*)$ yet to be specified.
    \item[2.] {\bf Noun argument prediction.} Here we predict which nouns $n^*$ are plausible novel arguments for a given verb-syntax frame $f$, operationalized as $p(n^*|f)$ yet to be specified.
\end{itemize}

We solve these inference problems by modeling the joint probability $p(n^*, f)$ for a query noun and a candidate verb-syntactic frame incrementally through time  as follows:
\begin{align}
p(n^*, f)^{(t)}  &= p(n^*|f)^{(t)}p(f)^{(t)} \label{joint} \\ 
                                        &= p(n^*|S(f)^{(t)})p(f)^{(t)}
\end{align}
Here we construct verb meaning based on its existing support nouns $S(f)^{(t)}$ at current time $t$. We infer the most probable verb-syntax usages for describing the query noun (Problem 1) as follows:
\begin{align}
    p(f|n^*) &= \frac{p(n^*, f)^{(t)}}{\sum_{f \in \mathcal{F}}p(n^*, f)^{(t)}} \\
             &= \frac{p(n^*|S(f)^{(t)})p(f)^{(t)}}{\sum_{f' \in \mathcal{F}}p(n^*|S(f')^{(t)})p(f')^{(t)}}
\end{align}
In the learning phase, we train our model incrementally at each time period $t$ by minimizing the log joint probability $p(n^*,f)^{(t)}$ in Equation~\ref{joint} for every frame $f$ and each of its query noun $n^* \in Q(f)^{(t)}$:
\begin{align}
    J = -\sum\limits_{f\in\mathcal{F}^{(t)}}\sum\limits_{n^*\in Q(f)^{(t)}}\log p(n^*, f)^{(t)}
\label{nll_loss}
\end{align}
  For each noun $n$, we consider a time-dependent hidden representation $\textbf{h}_{n}^{(t)} \in \mathbb{R}^{M}$ derived from different sources of knowledge (specified in Section~\ref{multimodal}). For the prior probability $p(f)^{(t)}$, we consider a frequency-based approach that computes the proportion for the number of unique noun arguments that a (verb) frame has been paired with and attested in a historical corpus:
 \begin{align}
     p(f)^{(t)} = \frac{|S(f)^{(t)}|}{\sum_{f'\in \mathcal{F}}|S(f')^{(t)}|}
 \end{align}
We formalize $p(n^*|f)$ (Problem 2), namely $p(n^*|N(f))^{(t)}$, by two classes of deep categorization models motivated by the literature on chaining, categorization, and memory-augmented learning. 

{\bf Deep prototype model (SFEM-DPM).} SFEM-DPM draws inspirations from prototypical network for few-shot learning \cite{snell2017prototypical} and is grounded in the prototype theory of categorization in cognitive psychology \cite{rosch1975cognitive}. The model computes a set of hidden representations for every support noun $n_s \in S(f)^{(t)}$, and takes the expected vector as a \textit{prototype} $\textbf{c}_f^{(t)}$ to represent $f$ at time $t$: 
\begin{align}
    \textbf{c}_f^{(t)} = \frac{1}{|S(f)^{(t)}|} \sum_{n_s \in S(f)^{(t)}} \textbf{h}_{n_s}^{(t)}
\end{align}
The likelihood of extending $n^*$ to $S(f)^{(t)}$ is then defined as a softmax distribution over $l_2$ distances $d(\cdot,\cdot)$ to the embedded prototype:
\begin{align}
    p(n^{*}|S(f)^{(t)}) = \frac{\exp{(-d(\textbf{h}_{n^*}^{(t)}, \textbf{c}_f^{(t)}))}}{\sum_{f'}\exp(-d(\textbf{h}_{n^*}^{(t)}, \textbf{c}_f^{(t)}))}
\end{align}
{\bf Deep exemplar model (SFEM-DEM).} In contrast to the prototype model, SFEM-DEM resembles the memory-augmented matching network in deep learning \cite{vinyals2016matching}, and formalizes the exemplar theory of categorization~\cite{nosofsky1986attention} and chaining-based category growth~\cite{habibi2020chaining}. Unlike DPM, this model depends on the $l_2$ distances between $n^*$ and every support noun:
\begin{align}
     \resizebox{\linewidth}{!}{$p(n^{*}|S(f)^{(t)})  
    = \frac{\sum\limits_{n_s \in S(f)^{(t)}}\exp{(-d(\textbf{h}_{n^*}^{(t)}, \textbf{h}_{n_s}^{(t)}))}}{\sum\limits_{f'}\sum\limits_{n'_s \in S(f')^{(t)}}\exp{(-d(\textbf{h}_{n^*}, \textbf{h}_{n'_s}^{(t)}))}}$}
\end{align}

\begin{table*}[t]
\resizebox{\textwidth}{!}{%
\begin{tabular}{@{}lllll@{}}
\toprule
\multirow{2}{*}{Decade} & \multicolumn{2}{c}{Verb syntactic frame}            & \multirow{2}{*}{Support noun} & \multirow{2}{*}{Query noun} \\ \cmidrule(lr){2-3}
                        & Predicate verb & Syntactic relation            &                                &                              \\ \cmidrule(r){1-1} \cmidrule(l){4-5} 
1900                    & drive          & direct object                 & horse, wheel, cart             & car, van                     \\
1950                    & work           & prepositional object via \it{as} & mechanic, carpenter, scientist & astronaut, programmer        \\
1980                    & store          & prepositional object via \it{in} & fridge, container, box         & supercomputer                \\ \bottomrule

\end{tabular}%
}
\caption{Sample entries from Google Syntactic Ngram  including verb syntactic frames,  support and query nouns.}
\label{ds_examples}
\end{table*}

\subsection{Multimodal  knowledge integration} \label{multimodal}

In addition to the probabilistic formulation, SFEM draws on structured knowledge including perceptual, conceptual, and linguistic cues to construct multimodal semantic representations $\textbf{h}_n^{(t)}$ introduced in Section~\ref{prob}. 

{\bf Perceptual knowledge.} We capture perceptual knowledge from image representations in the large, taxonomically organized ImageNet database~\cite{deng2009imagenet}. For each noun $n$, we randomly sample a collection of 64 images from the union of all ImageNet synsets that contains $n$, and encode the images through the VGG-19 convolutional neural network~\cite{vgg} by extracting the output vector from the last fully connected layer after all convolutions (see similar procedures also in \citealp{pinto2021computational}). We then average the encoded images to a
mean vector $\textbf{x}_p(n) \in \mathbb{R}^{1000}$ as the perceptual representation of $n$. 

{\bf Conceptual knowledge.} To capture conceptual knowledge beyond perceptual information (e.g., attributes and functions), we extract information from the ConceptNet knowledge graph~\cite{speer2017conceptnet}, which connects concepts in a network structure via different types of relations as edges. This graph reflects commonsense knowledge of a concept (noun) such as its functional role (e.g., a car IS\_USED\_FOR \textit{transportation}), taxonomic information (e.g., a car IS\_A \textit{vehicle}), or attributes (e.g., a car HAS\_A \textit{wheel}). Since the concepts and their relations may change over time, we prepare a diachronic slice of the ConceptNet graph at each time $t$ by removing all words with frequency up to $t$ in a reference historical text corpus (see Section~\ref{data} for details) under a threshold $k_\text{c}$ which we set to be 10. We then compute embeddings for the remaining concepts following methods recommended in the original study by \citet{speer2017conceptnet}. In particular, we perform singular value decomposition (SVD) on the positive pointwise mutual information matrix $\textbf{M}_G^{(t)}$ of the ConceptNet $G^{(t)}$ truncated at time $t$, and combine the top 300 dimensions (with largest singular values) of the term and context matrix symmetrically into a concept embedding matrix. Each row of the resulting row matrix of SVD will therefore serves as the conceptual embedding $\textbf{x}_c(n)^{(t)} \in \mathbb{R}^{300}$ for its corresponding noun.

{\bf Linguistic knowledge.} For linguistic knowledge, we take the HistWords diachronic word embeddings $\textbf{x}_{l}^{(t)} \in \mathbb{R}^{300}$ pre-trained on the Google N-Grams English corpora to represent linguistic meaning of each noun at decade $t$ \cite{hamilton2016diachronic}. 

{\bf Knowledge integration.} To construct a unified representation that incorporates knowledge from different modalities, we take the mean of the unimodal representations described into a joint  vector $\textbf{x}_{n} \in \mathbb{R}^{300}$, and then apply an integration function $g: \mathbb{R}^{300} \rightarrow \mathbb{R}^M$ parameterized by a feedforward neural network to get the multimodal word representation $h_n^{(t)}$.\footnote{For ImageNet embeddings, we apply a linear transformation to project each $\textbf{x}_{p}^{(t)}$ into $\mathbb{R}^{300}$ so that all unimodal representations are 300-d vectors before taking the means.} Our framework allows flexible combinations of the three modalities introduced, e.g., a full model would utilizes all three types of knowledge, while a linguistic-only baseline will directly take HistWords embeddings $\textbf{x}_{l}^{(t)}$ as inputs of the integration network.

\section{Historical noun-verb compositions} \label{data}

To evaluate our framework, we collected a large dataset of historical noun-verb compositions derived from the Google Syntactic N-grams (GSN) English corpus \cite{lin2012syntactic} from 1850 to 2000. Specifically, we collected verb-noun-relation triples $(n,v,r)^{(t)}$ that co-occur in the ENGALL subcorpus of GSN over the 150 years. We focused on working with common usages and pruned rare cases under the following criteria: 1) a noun $n$ should have at least $\theta_{p} = 64$ image representations in ImageNet,  $\theta_{c} = 10$ edges in the contemporary ConceptNet network, and $\theta_{n} = 15,000$ counts (with POS tag as nouns) in GSN over the 150-year period; 2) a verb $v$ should have at least $\theta_{v} = 15,000$ counts in GSN. To facilitate feasible model learning, we consider the top-20 most common syntactic relations in GSN, including direct object, direct subject, and relations concerning  prepositional objects.

We binned the raw co-occurrence counts by decade $\Delta=10$. At each decade, we define emerging query nouns $n^*$ for a given verb frame $f$ if their number of co-occurrences with $f$ up to time $t$ falls below a threshold $\theta_{q}$, while the number of co-occurrences with $f$ up to time $t+\Delta$ is above $\theta_{q}$ (i.e., an emergent use that conventionalizes). We define support nouns as those that co-occurred with $f$ for more than $\theta_{s}$ times before $t$. We found that $\theta_{q}=10$ and $\theta_{s}=100$ are reasonable choices. This preprocessing pipeline  yielded a total  of 10,349 verb-syntactic frames over 15 decades, where each frame class has at least 1 novel query noun and 4 existing support nouns. Table \ref{ds_examples} shows sample entries of data which we make publicly available.\footnote{Data and code are deposited here: \url{https://github.com/jadeleiyu/frame_extension}}

\section{Evaluation and results}

We first describe the  details of SFEM implementation and diachronic evaluation. We then provide an in-depth analysis on the  multimodal knowledge and chaining mechanisms in  verb frame extension.

\subsection{Details of model implementation}
We implemented the integration network $g(\cdot)$ of SFEM as a three-layer feedforward neural network with an output dimension $M=100$, and keep parameters and embeddings in other modules fixed during learning.\footnote{See Appendix A for additional implementation details.} At each decade, we randomly sample $70\%$ of the query nouns with their associated verb-syntactic pairs as training data, and take the remaining examples for model testing such that there is no overlap in the query nouns between training and testing. We trained models on the negative log-likelihood loss defined in Equation \ref{nll_loss} at each decade. To examine how multimodal knowledge contributes to temporal prediction of novel language use, we trained 5 DEM and 5 DPM models using information from different modalities.

\begin{figure*}[h]
\centering
\includegraphics[width=\textwidth,height=8cm]{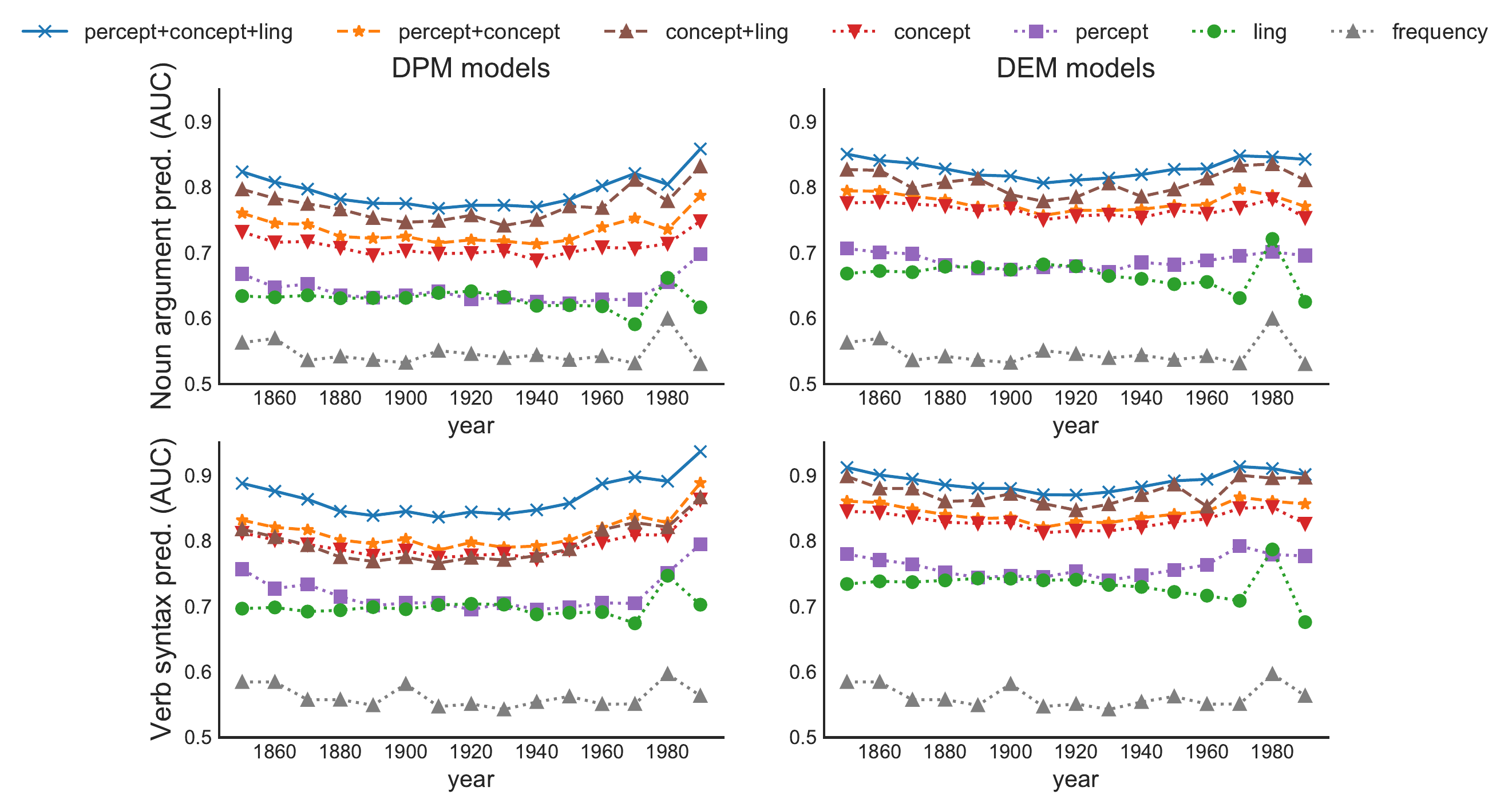}
\caption{Area-under-curves of SFEM and baseline models from 1850s to 1990s. Top row: AUCs of predicting extended syntax frames for query nouns. Bottom row: AUCs of predicting extended nouns for query verb frames.}
\label{aucs_by_decade}
\end{figure*}

\begin{table*}[h]
\resizebox{\linewidth}{!}{%
\begin{tabular}{@{}lllllll@{}}
\toprule
\multicolumn{1}{c}{\multirow{2}{*}{Model}} & \multicolumn{3}{c}{AUC -- verb syntax prediction} & \multicolumn{3}{l}{AUC -- noun argument prediction} \\ \cmidrule(l){2-7} 
\multicolumn{1}{c}{}                       & novel items    & existing items    & combined      & novel items    & existing items    & combined    \\ \cmidrule(r){1-1}
DPM (linguistics)               & 0.642        & 0.690             & 0.681    & 0.641        & 0.653             & 0.650  \\
DPM (perceptual)                    & 0.632        & 0.666             & 0.657    & 0.650        & 0.624             & 0.629   \\
DPM (conceptual)                & 0.772        & 0.722             & 0.733    & 0.727        & 0.705             & 0.711   \\
DPM (perceptual+conceptual)              & 0.809        & 0.754             & 0.767    & 0.725        & 0.719             & 0.721   \\ 
DPM (perceptual+linguistics)              & 0.645        & 0.669             & 0.661    & 0.655        & 0.669             & 0.665   \\ 
DPM (conceptual+linguistics)              & 0.753        & 0.774             & 0.766    & 0.776        & 0.768             & 0.770   \\ 
DPM (perceptual+conceptual+linguistics)              & 0.848        & 0.810             & 0.815    & 0.799        & 0.786             & 0.788 \\\hline
DEM (linguistics)                & 0.652        & 0.690             & 0.686    & 0.641        & 0.625             & 0.632   \\
DEM (perceptual)                     & 0.737        & 0.674             & 0.684    & 0.659        & 0.650             & 0.655   \\
DEM (conceptual)                 & 0.854        & 0.784             & 0.788    & 0.736        & 0.724             & 0.729   \\
DEM (perceptual+conceptual)               & 0.858        & 0.792             & 0.797    & 0.750        & 0.744             & 0.748   \\ 
DEM (perceptual+linguistics)               & 0.712        & 0.759             & 0.753    & 0.698        & 0.710             & 0.708   \\ 
DEM (conceptual+linguistics)               & 0.902        & 0.866             & 0.870    & 0.837       & {\bf 0.822}          & 0.824   \\ 
DEM (perceptual+conceptual+linguistics)              & {\bf 0.919}        & {\bf 0.872}             & {\bf 0.878}    & {\bf 0.856}        & 0.820             & {\bf 0.827} \\\hline
Baseline (frequency)                       & 0.573        & 0.573             & 0.573    & 0.536        & 0.536             & 0.536   \\
Baseline (random)                          & 0.500        & 0.500             & 0.500    & 0.500        & 0.500             & 0.500   \\\bottomrule
\end{tabular}%
}
\caption{Mean model AUC scores of verb syntax and noun argument predictions from 1850s to 1990s. }
\label{mean_aucs_table}
\end{table*}

\begin{table*}[h!]
\resizebox{\textwidth}{!}{%
\begin{tabular}{llll}
\hline
Query noun    & Decade & Predicted frames (linguistic-only DEM)  & Predicted frames (tri-modal DEM)                                                                                             \\ \hline
telephone     & 1860   & \begin{tabular}[c]{@{}l@{}}roll-nsubj, load-dobj, \\ play-pobj\_prep.on\end{tabular}                              & \begin{tabular}[c]{@{}l@{}}purchase\_dobj, pick-dobj, \\ remain-pobj\_prep.on\end{tabular}                                           \\\hline
microorganism & 1900   & \begin{tabular}[c]{@{}l@{}}decorate-pobj\_prep.with, \\ play-pobj\_prep.on, spread-dobj\end{tabular}              & \begin{tabular}[c]{@{}l@{}}feed-pobj\_prep.on, \\ mix-pobj\_prep.with, breed-dobj\end{tabular}                                       \\\hline
airplane      & 1930   & \begin{tabular}[c]{@{}l@{}}load-dobj, mount-pobj\_prep.on, \\ mount-dobj, blow-dobj, roll-nsubj\end{tabular}      & \begin{tabular}[c]{@{}l@{}}fly-dobj,  approach-nsubj,\\  drive-dobj, stop-nsubj\end{tabular}                                         \\\hline
astronaut     & 1950   & \begin{tabular}[c]{@{}l@{}}spin-dobj,work-pobj\_prep.in,\\  emerge-pobj\_prep.from\end{tabular}                   & \begin{tabular}[c]{@{}l@{}}work-pobj\_prep.as, talk-pobj\_prep.to, \\ lead-pobj\_prep.by\end{tabular}                                \\\hline
computer & 1970   & \begin{tabular}[c]{@{}l@{}}purchase-dobj, fix-dobj, \\ generate-dobj, write-pobj\_prep.to\end{tabular} & \begin{tabular}[c]{@{}l@{}}store-pobj\_prep.in, move-pobj\_prep.into, \\ display-pobj\_prep.on, implement-pobj\_prep.in\end{tabular} \\ \hline

\end{tabular}%
}
\caption{Example predictions of novel verb-noun compositions from the full tri-imodal and linguistic-only models.}
\label{prediction_examples_table}
\end{table*}

\begin{table}[]
\resizebox{\columnwidth}{!}{%
\begin{tabular}{@{}ll@{}}
\toprule
Ablated modality & Most affected compositions                                                                                                                                     \\ \midrule
Language         & \begin{tabular}[c]{@{}l@{}}buy a monitor, find a disk (*), \\ a resident dies, \\ specialized in nutrition (*),\\ point to the window\end{tabular}             \\ \midrule
Percept          & \begin{tabular}[c]{@{}l@{}}an airplane rolls, \\ talk to an entrepreneur (*),\\ the tree stands, the doctor says,\\ topped with nutella (*)\end{tabular}       \\ \midrule
Concept          & \begin{tabular}[c]{@{}l@{}}perform in the film (*), \\ work as a programmer (*),\\ work for the newspaper, \\ expand the market, kill the process\end{tabular} \\ \bottomrule
\end{tabular}%
}
\caption{Top-4 ground-truth compositions with most prominent drops in joint probability $p(f,n)$ after ablation of one modality of knowledge from SFEM. Phrases marked with `*' include novel query nouns.}
\label{top5tab}
\end{table}

\subsection{Evaluation against historical data}

We test our models on both verb syntax and noun argument predictive tasks with the goals of assessing 1) the contributions of multimodal knowledge, and 2) the two alternative mechanisms of chaining. We also consider baseline models that do not implement chaining-based mechanisms: a frequency baseline that predicts by count in GSN up to time $t$, and a random guesser. We evaluate model performance via standard receiver operating characteristics (ROC) curves that reveal cumulative precision of models in their  top $m$ predictions. We compute the standard area-under-curve (AUC) statistics for the ROC curves to get the mean precision over all values of $m$ from 1 to the candidate set size. Figure \ref{aucs_by_decade} summarizes the results over the 150 years. We observe that 1) all multimodal models perform better than their uni-/bi-modal counterparts, and 2) the exemplar-based model performs dominantly better than the prototype-based counterpart, and both outperform the baseline models without chaining. In particular, a tri-modal deep exemplar model that incorporates knowledge from all three modalities achieves the best overall performance. These results provide strong  support that verb frame extension depends on multimodal knowledge and an exemplar-based  chaining.

To further assess how the models perform in predicting emerging verb extension toward both novel and existing nouns, we report separate mean AUC scores for these predictive cases where query nouns are either completely novel (i.e., zero token frequencies) or established. (i.e., above-zero frequencies) at the time of prediction.  Table \ref{mean_aucs_table} summarizes these results and shows that model performances are similar under four predictive cases. For prediction with novel query nouns, it is not surprising that linguistic-only models fail due to the unavailability of linguistic mentions. However, for prediction with established query nouns, the superiority of multimodal SFEMs is still prominent suggesting that our framework captures general principles in verb frame extension (and not just for predicting verb extension toward novel nouns).

Table \ref{prediction_examples_table} compares sample verb syntax predictions made by the full and linguistic-only DEM models that cover a diverse range of  concepts including inventions (e.g., airplane),  discoveries (e.g., microorganism), and  occupations (e.g. astronaut). We observe that the full model typically constructs reasonable predicate verbs that reflect salient features of the query noun (e.g., \textit{cars} are vehicles that are \textit{drive}-able). In contrast, the linguistic-only model often predicts verbs that are either overly generic (e.g., \textit{purchase} a telephone) or  nonsensical.

\subsection{Model analysis and interpretation}

We provide further analyses and interpret why both multimodality and chaining mechanisms are fundamental to predicting emergent verb compositions.

\subsubsection{The function of multimodal knowledge}

To  understand the function of multimodal knowledge, we compute, for each modality, the top-4 verb compositions that were most degraded in joint probability $p(f,n)$ after ablating a knowledge modality from the full tri-modal SFEM (see Table \ref{top5tab}). A drop in $p(f,n)$ indicates reliance on multimodality in prediction. We found that linguistic knowledge  helps the model identify some general properties that are absent in the other cues (e.g., a \textit{monitor} is \textit{buy}-able). Importantly, for the two extra-linguistic knowledge modalities, we observe that visual-perceptual knowledge helps predict many imaged-based metaphors, including ``the airplane rolls" (i.e., based on  common shapes of airplanes) and ``the tree stands" (based on verticality of trees). On the other hand, conceptual knowledge predicts  cases of logical metonymy (e.g., ``work for the newspaper'') and conceptual metaphor (e.g., ``kill the process''). These examples suggest that multimodality serves to ground and embody SFEM with commonsense knowledge that constructs novel verb compositions for not only literal language use, but also non-literal or figurative language use that is extensively discussed in the psycholinguistics literature \cite{lakoff1982experiential, radden1999towards, gibbs2004metaphor}.

\begin{figure}[]
\centering
\includegraphics[width=1\linewidth]{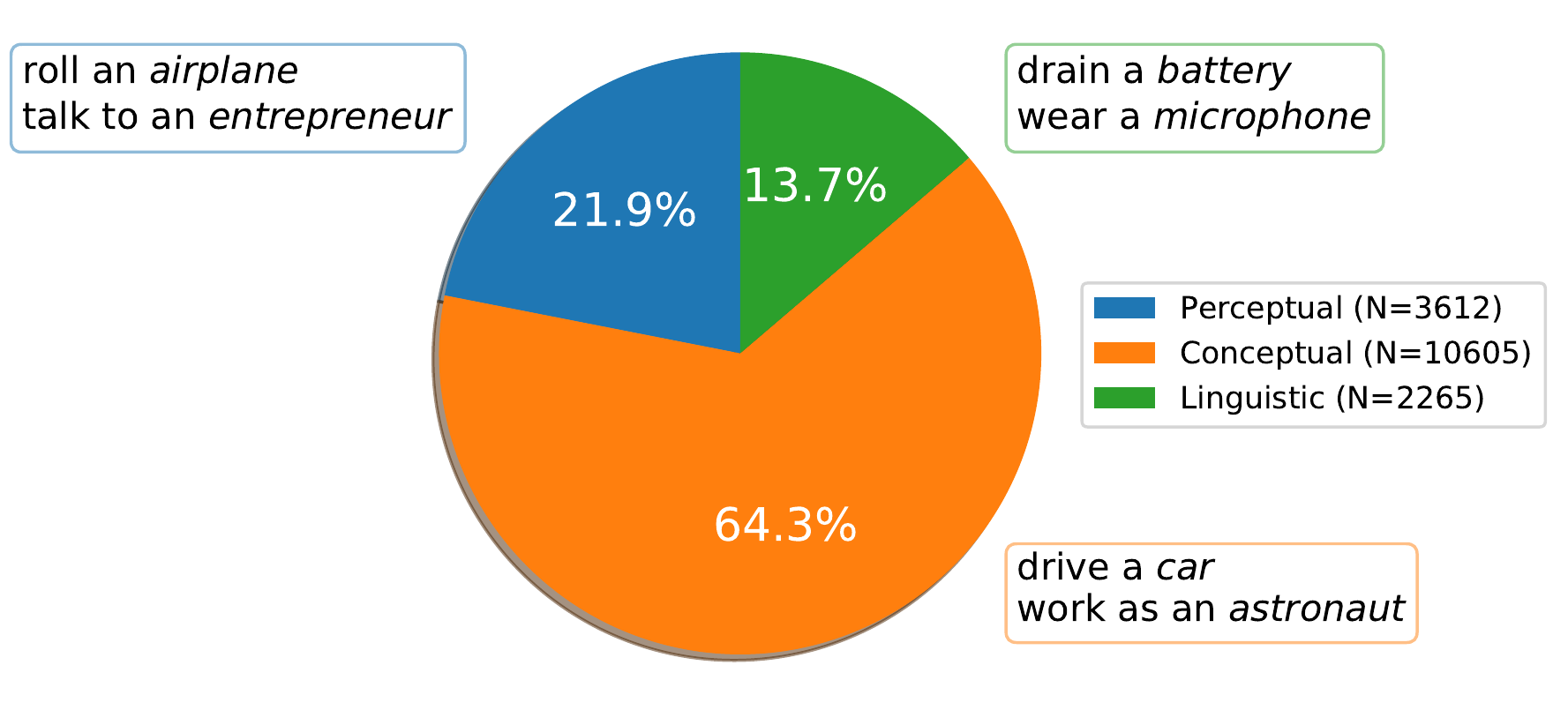}
\caption{Percentage breakdown of the three modalities in model prediction, with annotated examples.}
\label{modes_pie}
\end{figure}

We also evaluate the contributions of the three modalities in model prediction by comparing the AUC scores from the three uni-modal DEMs. Figure \ref{modes_pie} shows the percentage breakdown of examples on which one of the modalities yields the highest score (i.e., contributes most to a reliable prediction). We observe that conceptual cues explain data the best in almost 2/3 of the cases, followed by perceptual and linguistic cues. These results suggest that while conceptual knowledge plays a dominant role in model prediction, all three modalities contain complementary information in predicting novel language use through time.

\subsubsection{General mechanisms of chaining}

\begin{figure*}[ht!]
\centering
\includegraphics[scale=.75]{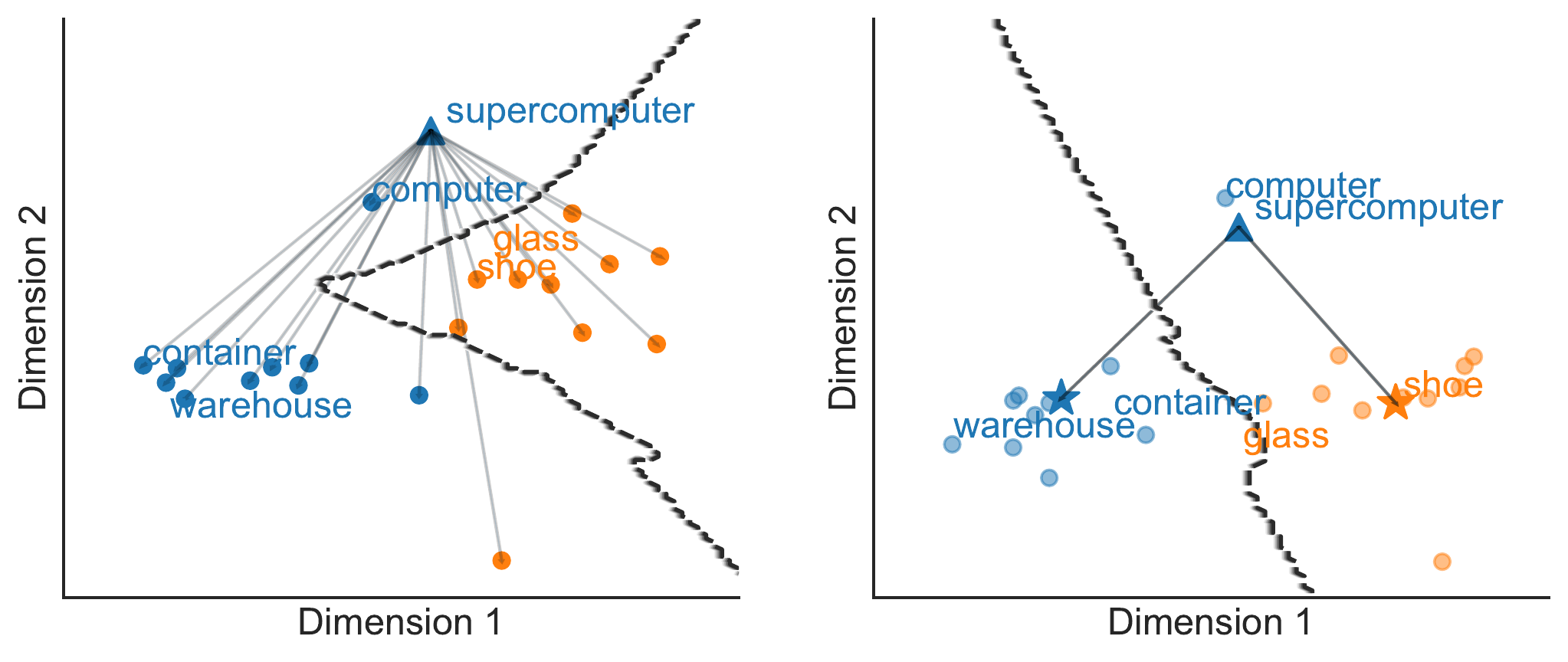}
\caption{Illustrations of two mechanisms of chaining (left: exemplar; right: prototype) in verb frame prediction for query \textit{supercomputer}. Nouns are PCA-projected in 2D, with categories color-coded and in dashed boundaries.}
\label{chaining_illustration}
\end{figure*}

We next analyze general mechanisms of chaining by focusing on understanding the superiority of exemplar-based chaining in SFEM. Figure \ref{chaining_illustration} illustrates the exemplar-based and prototype-based processes of chaining with the example verb frame prediction for the noun ``supercomputer". For simplicity, we only show two competing frames ``to store in a \_\_\_'' and ``to wear a \_\_\_''. In this case, the query noun is semantically distant to most of the prototypical support nouns in both categories, and is slightly closer to the centroid of the ``wear'' class than to that of the ``store'' class. The prototype model would then predict the incorrect composition ``to wear a supercomputer''. In contrast, the exemplar model is more sensitive to the semantic neighborhood profile of the query noun and the aprototypical support noun ``computer'' of the ``store in \_\_\_'' class, and it therefore  correctly predicts that ``supercomputer'' is more likely to be predicated by ``to store in''. Our discovery that the exemplar-based chaining accounts for verb composition through time mirrors existing findings on similar mechanisms of chaining in the extensions of numeral classifiers~\cite{habibi2020chaining} and adjectives~\cite{grewal2020chaining}, and together they suggest a general cognitive mechanism may underlie historical linguistic innovation.


\section{Conclusion}
We have presented a probabilistic framework  for characterizing the process of syntactic frame extension in which verbs extend their referential range toward novel and existing nouns over time. Our results suggest that language users rely on extra-linguistic knowledge from percept and concept to construct new linguistic compositions via a process of exemplar-based chaining. Our work creates a novel approach to diachronic compositionality and strengthens the link between multimodal semantics and cognitive linguistic theories of categorization. 

\section*{Acknowledgments}
We would like to thank Graeme Hirst, Michael Hahn, and Suzanne Stevenson for their feedback on the manuscript, and members of the Cognitive Lexicon Laboratory at the University of Toronto for helpful suggestions. We also thank the anonymous reviewers for their constructive comments. This work was supported by a NSERC Discovery Grant RGPIN-2018-05872, a SSHRC Insight Grant \#435190272, and an Ontario ERA Award to YX.

\bibliography{anthology,custom}
\bibliographystyle{acl_natbib}

\clearpage

\appendix

\section{Additional details of SFEM implementation}

We implemented the integration network $g(\cdot)$ as a three-layer feedforward neural network using PyTorch, where each layer has a dimension of $300, 200$ and $100$ respectively. For models that incorporates less than three modalities, we replace the missing embeddings with a zero vector when computing the mean vectors before knowledge integration. 

During training, except for network weights in $g(\cdot)$, we keep parameters in every modules (i.e., the VGG-19 encoder and every unimodal embedding) constant, and optimize SFEM by minimzing the negative log-likelihood loss function specified in Equation \ref{nll_loss} via stochastic gradient descent (SGD). 

Each training batch consists of $B=64$ syntactic frames with their associated query and support nouns. We train each model for 200 epochs and save the configuration that achieves the highest validation accuracy for our evaluation described in Section 5. 
\label{appendixA}

\end{document}